\documentclass[11pt]{article}

\usepackage[final]{acl}

\usepackage[utf8]{inputenc}
\usepackage[T1]{fontenc}
\usepackage{times}
\usepackage{latexsym}
\usepackage{url}
\usepackage{booktabs}
\usepackage{multirow}
\usepackage{makecell}
\usepackage{amsfonts}
\usepackage{amsmath}
\usepackage{amssymb}
\usepackage{nicefrac}
\usepackage{microtype}
\usepackage{graphicx}

\title{LatentPress: Context Compression Beyond Text and Vision}

\author{Yuanda Xu\thanks{Equal contribution}\, \thanks{Correspondence to \texttt{yuanda@math.princeton.edu}} \quad
    Zhengze Zhou\footnotemark[1]$^{2}$\quad
    Hejian Sang\footnotemark[1] \quad
    Xinchen Du$^{3}$ \quad
    Baiyang Dai \quad
    Sen Na$^{3}$ \\
    \vspace{0.2cm}
    $^{2}$Cornell University \quad
    $^{3}$Georgia Institute of Technology
}

\begin{document}

\maketitle

\begin{abstract}
Compressed context is usually carried as human-readable text or as rendered images that must be decoded, even when its consumer is a language model. We introduce \textbf{LatentPress}, which writes conversational histories and long documents into a third representation: continuous memory tokens that a frozen decoder reads directly through its input-embedding interface, with no text reconstruction at inference. A small reader-matched writer compresses $4$--$16\times$ while training only an adapter ($4.2$M--$26.2$M parameters, $0.18$--$0.25\%$ of the decoder). On LongMemEval, LatentPress reaches $0.504$ accuracy at $7.70\times$ compression versus $0.490$ for uncompressed evidence,
outperforming text summaries ($0.184$) and OCR-based compression ($0.426\!\to\!0.312$). On LongBench-QA, in-domain writers match or exceed
raw-context reading at $4$--$8\times$ compression, while $16\times$ trails raw. Writing takes $43$\,ms per conversation, roughly an order of magnitude faster than text summarization or OCR reconstruction, and reading is $5$--$9\times$ faster than raw context or cached OCR. We validate the interface under two transfer settings, zero-shot from UltraChat to LongMemEval memory QA and from LongMemEval-derived QA to unseen LongBench document domains, establishing direct soft tokens as a practical machine-facing context interface beyond text and vision. 
The implementation of the experiments could be found at: https://github.com/HJSang/LatentPress
\end{abstract}

\section{Introduction}

\begin{figure*}[t]
  \centering
  \includegraphics[width=\linewidth]{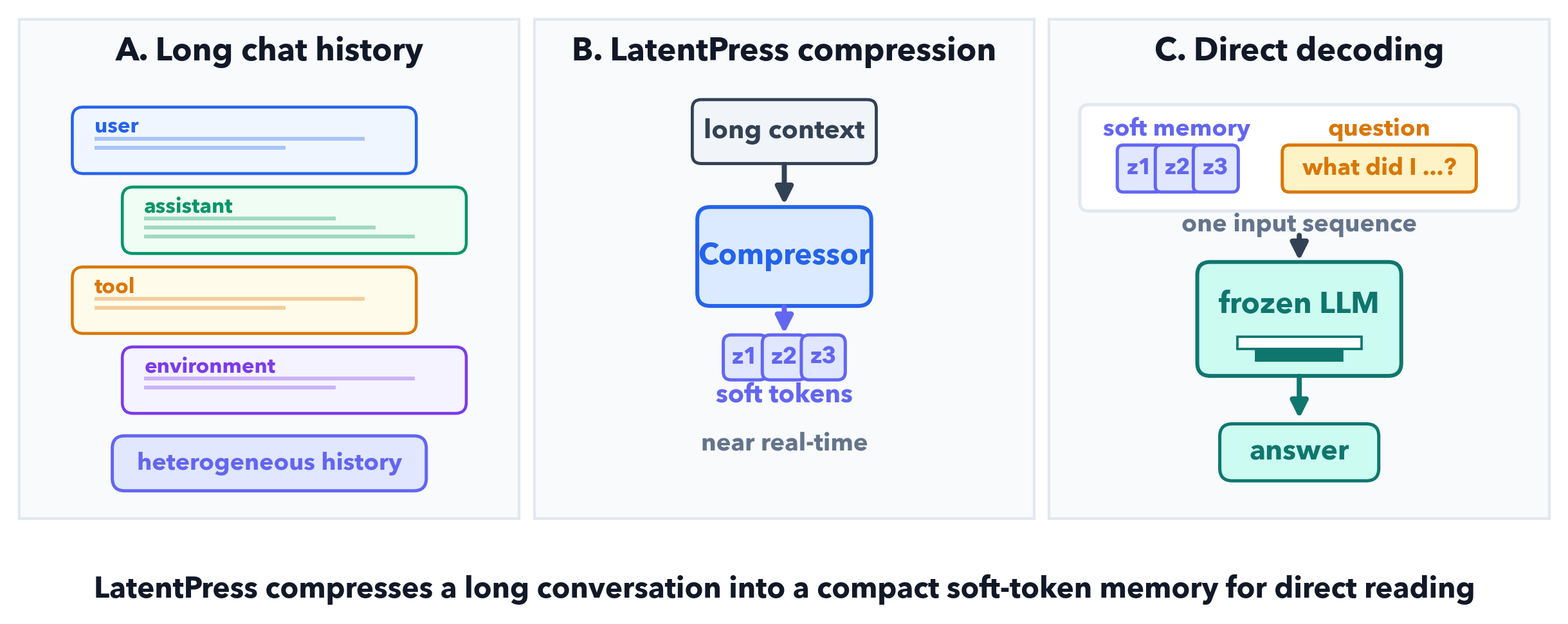}
  \caption{\textbf{LatentPress overview, shown for conversational context.}
  \textbf{(A)} A long, heterogeneous history containing segments with different
  information value. \textbf{(B)} LatentPress compresses the context into a short
  sequence of continuous soft tokens in a single, near-real-time forward pass.
  \textbf{(C)} The frozen LLM reads one concatenated sequence (the soft-token
  context followed by the question) and decodes the answer \emph{directly},
  with no text-reconstruction step. We also evaluate the same direct-read
  interface on long documents as a generalization beyond conversational memory.}
  \label{fig:overview}
\end{figure*}

Long-running assistants and agents accumulate more history than they can afford
to reread. A deployment trace may hold instructions, dialogue, plans, tool calls,
observations, and environment feedback, yet a later decision often depends on
only a small part of it. The same pressure appears whenever a language model must
read long documents to answer a question. In both cases, the default
machine-facing interface remains discrete text: systems retrieve text, summarize
text, prune text, or reconstruct text from another modality before a language
model can use it. Text is convenient for people and interoperable across systems,
but a model need not require its stored or compressed context to be
human-readable. This motivates a more direct question: can long context be
written into a compact continuous representation that a frozen language model
reads without first recovering the text?

We study this question at the representation layer. We separate context use into
\textsc{Write}, which maps text to a compact state, and \textsc{Read}, which
supplies that state to a frozen decoder for downstream QA. This abstraction
covers both conversational histories and long documents. It does not attempt to
replace retrieval, reflection, update policies, or conflict resolution in a
complete memory system; instead, it asks what representation should cross the
boundary between stored context and the model that consumes it.

We introduce \textbf{LatentPress}, a direct-read soft-token interface
(Figure~\ref{fig:overview}). A reader-matched writer reuses two frozen decoder
layers together with a small trainable adapter to map text segments into
continuous vectors. These vectors enter the frozen decoder through its
input-embedding interface, followed by the question.

Making this interface useful requires two practical choices: how aggressively to
compress each segment and what supervision teaches the writer to retain. The
contribution we emphasize is the interface itself, not the compression schedule:
for the per-segment rate we simply exploit whatever structure the input already
exposes, and we treat \emph{where} to spend the compression budget as a
hand-specified heuristic rather than a learned component. Since token positions
differ widely in how much they contribute~\citep{xu2026tip}, conversational turns
follow a simple structure-based schedule while unstructured long documents use a
single uniform rate; learning this allocation automatically is a direction we
leave to future work. For documents we additionally study how cross-domain and
in-domain QA supervision affects the compressed reader.

\paragraph{How LatentPress differs from prior compression.}
Compressing context into continuous vectors is an established family, so we state
up front what LatentPress changes (Table~\ref{tab:positioning}). Gist,
AutoCompressor, and ICAE train or adapt an LLM-scale reader or encoder, whereas
LatentPress leaves the downstream decoder entirely frozen and trains only a small
reader-matched adapter ($0.18$--$0.25\%$ of decoder parameters). Unlike ICAE and
visual compression, its vectors are consumed directly at the decoder's
input-embedding layer, with no text reconstruction at inference. xRAG also
freezes the reader, but compresses one independently retrieved passage into a
single token; LatentPress instead writes multi-turn histories and whole long
documents, and can assign different rates to structured segments. The resulting
distinction is not soft tokens alone, but a lightweight \textsc{Write}/\textsc{Read}
interface that combines a frozen reader, direct soft-token consumption, and
variable-length context compression. We discuss the closest mechanisms and use
cases in the Related Work.

\begin{table*}[t]
  \centering
  \caption{\textbf{Positioning among continuous-vector context compression
  methods.} ``FT'' is full fine-tuning; ``autoenc.'' decodes memory vectors back
  to text before answering.}
  \label{tab:positioning}
  \resizebox{\textwidth}{!}{%
  \begin{tabular}{lccccc}
    \toprule
    Method & What is trained & Trainable & Representation & Reconstruct at & Rate \\
     & (mechanism) & scale & space & inference? & \\
    \midrule
    Gist~\citep{mu2023gist} & whole decoder (FT, masked attn.) & decoder-scale & KV-cache & no & fixed \\
    AutoCompressor~\citep{chevalier2023autocompressor} & LLM (recursive summary) & LLM-scale & input (summary) & no & uniform \\
    ICAE~\citep{ge2024icae} & LLM encoder (LoRA) & LLM-scale (LoRA) & input slots & \textbf{yes} (autoenc.) & uniform \\
    xRAG~\citep{cheng2024xrag} & projector only (LLM frozen) & small projector & input (1 token) & no & single-token \\
    DeepSeek-OCR~\citep{wei2025deepseekocr} & vision model & vision-model-scale & image$\to$text & \textbf{yes} (OCR) & resolution \\
    \midrule
    \textbf{LatentPress (ours)} & \textbf{Only adapter} & \textbf{$0.18$--$0.25\%$} & input-embedding & \textbf{no} & \textbf{variable, role-based} \\
     & \textbf{(decoder frozen)} & & & & \\
    \bottomrule
  \end{tabular}%
  }
\end{table*}

The experiments ask whether this interface is practical along four axes:
accuracy, write cost, read cost, and trainable footprint. LongMemEval tests the
accuracy and transfer behavior for conversational memory, where a writer trained
on generic UltraChat conversations transfers to unseen memory-QA labels across
three frozen readers ($0.48$--$0.50$ accuracy at $4.6$--$7.7\times$
compression). LongBench-QA~\citep{bai2024longbench} removes the role structure
and tests the same interface on long documents, both cross-domain and after
in-domain task adaptation, where in-domain compressed readers match or exceed
their raw-context baselines at mild compression while both transfer settings
degrade at the most aggressive rate. The efficiency section then
measures the two latency axes directly: encoded-token generation ($43$\,ms per
conversation) and warm-loaded inference from the compressed prefix
($5$--$9\times$ faster), using only a small trainable writer
($4.2$M--$26.2$M parameters).

\section{LatentPress}
\label{sec:method}

LatentPress is designed so that the expensive object, the downstream decoder, never
changes. This section defines the direct-read soft-token interface, the
reader-matched writer, and the two choices that determine what reaches the frozen
reader: the compression rate for each segment and the supervision used to train
the writer.

\subsection{Direct-read soft context}
Let a context $x=(x_1,\ldots,x_T)$ be a sequence of segments, such as dialogue
turns or document chunks. A frozen decoder $f_\theta$ answers a question $q$ from
a compact representation of this context. A small trainable writer maps $x$ to a
short sequence of continuous vectors $m$ that the decoder reads directly through
its input-embedding interface, followed by the embedded question:
\begin{equation}
  m = \textsc{Write}_{\phi}(x;\pi), \qquad
  y = f_\theta\big([m;\mathrm{emb}(q)]\big).
\end{equation}
Here $\phi$ denotes the writer parameters and $\pi$ specifies the compression
rate for each segment. For each position $i$, the writer fuses the literal input
embedding $E_i$ with a context-aware abstraction $c_i$ of it,
\begin{equation}
  h_i = H(E_i, c_i),
\end{equation}
where the general framework permits $H$ to be a learned, importance-weighted
fusion of literal and contextual features~\citep{srivastava2015highway,cho2014gru}.
In our lightweight instantiation, a small linear adapter maps the contextual
features into the reader's embedding space, where they are combined with the
original token embeddings before pooling; learned token-wise fusion is left to
future work. The resulting $h_i$ are pooled into a shorter
sequence of soft tokens that live in the reader's embedding space and are
injected into $f_\theta$ without changing any decoder weights. Only a lightweight
writer is trained, and because its soft tokens are tied to a specific reader we
train one writer per reader in the cross-reader experiments. Unlike
reconstruction-based interfaces, LatentPress never decodes the vectors back to text
at inference time, so writing is a single forward pass.

\subsection{Choosing compression rates}
The rule $\pi=(k_1,\ldots,k_T)$ determines how many neighboring token positions
are pooled into each soft token. We deliberately keep this rule simple and
hand-specified, since our aim is to test the direct-read interface rather than to
optimize the compression schedule; learning $\pi$ per segment is a direction we
leave to future work (Section~\ref{sec:future}). We study two such fixed rules.
\emph{Uniform} pooling sets $k_i=k$ for every segment; this is the document
configuration and the uniform dialogue comparison. A simple \emph{role-based}
schedule instead uses known input structure to vary $k_i$ across segments: for
conversational memory we set $k_i=k_{r_i}$ according to the turn role, with
$k_{\text{user}}=1$ and $k_{\text{assistant}}\in\{8,16,32\}$, so user turns bypass
the writer and retain their raw token embeddings while assistant turns are
encoded and pooled. The resulting conversation-level compression ratio emerges
from the role and length mixture rather than being a preset global rate.

\subsection{Small writer and bottleneck supervision}
The trainable footprint is intentionally small. Only the writer head is trained,
and its size is backbone-dependent:
$12.849$M parameters for Qwen2.5-7B, $16.781$M for Qwen3-8B, $4.196$M for
Qwen3-1.7B, and $26.220$M for the Qwen2.5-14B reader used in LongBench
experiments. The borrowed reader layers and the entire decoder are frozen.
We consider two sources of supervision. For generic representation learning, we
minimize
\begin{equation}
  \mathcal{L}(\phi)
  = \mathcal{L}_{\mathrm{rec}}
  + \lambda\,\mathcal{L}_{\mathrm{fkl}},
  \label{eq:training}
\end{equation}
where, for a target sequence $y=(y_1,\ldots,y_N)$,
\begin{align}
  \mathcal{L}_{\mathrm{rec}}
  &= -\frac{1}{N}\sum_{t=1}^{N}\log p_{\mathrm{comp},t}(y_t), \\
  \mathcal{L}_{\mathrm{fkl}}
  &= \frac{1}{N}\sum_{t=1}^{N}
  \operatorname{KL}\!\left(p_{\mathrm{full},t}\,\|\,
  p_{\mathrm{comp},t}\right).
\end{align}
Here $p_{\mathrm{full},t}$ and $p_{\mathrm{comp},t}$ are the frozen decoder's
teacher-forced next-token distributions given the full and compressed context,
respectively. The reconstruction term trains the compressed context to recover
the target tokens, while the forward-KL term distills the full-context behavior
into the writer, analogous in spirit to shortening a model's own reasoning
through self-distillation~\citep{sang2026crisp}. We use $\lambda=1.0$. The term \emph{reconstruction-free}
describes the inference interface, not this training signal: LatentPress never
reconstructs text before answering at evaluation time. For task adaptation, we
train the same writer on QA examples from either a
different domain or the target-domain training split, exposing the bottleneck to
the information demands of downstream reading. In both cases the writer and
compression rule change what reaches the reader, while $f_\theta$ remains frozen.

Below, the LongMemEval writer learns generic dialogue representations on
UltraChat and transfers zero-shot with role information, while the LongBench-QA
experiments use uniform compression and vary the QA supervision source.

\section{Conversational Memory}
\label{sec:longmemeval}

Conversational memory tests the first accuracy claim: a compressed soft-token
history can preserve the answer-relevant information that a frozen reader needs.
Each frozen decoder uses its own representation-matched writer, trained on
UltraChat and evaluated on unseen LongMemEval memory-QA conversations. We report
task accuracy alongside the ratio of original text tokens to injected vectors.

\begin{figure*}[t]
  \centering
  \includegraphics[width=\linewidth]{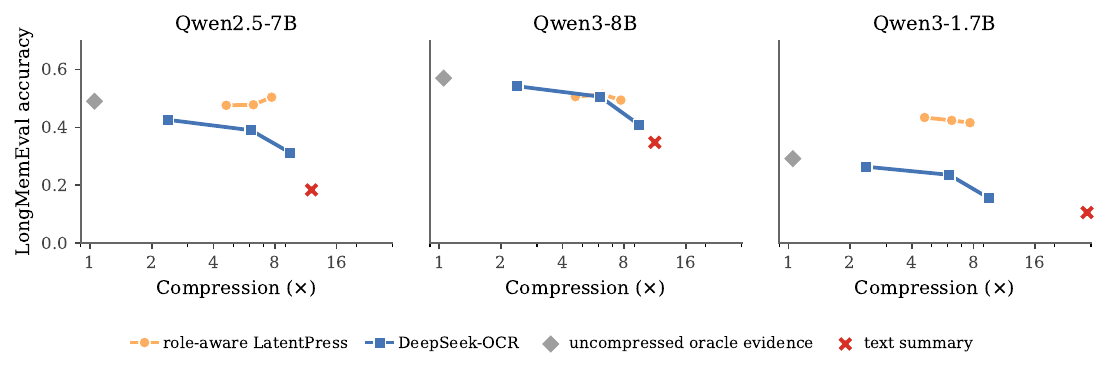}
  \caption{\textbf{LongMemEval accuracy--compression frontiers.}
  Role-aware LatentPress (orange) stays stable across compression rates on all three
  readers. Its relationship to the uncompressed oracle-evidence baseline (gray
  diamond) is reader-dependent: role-aware matches raw on Qwen2.5-7B, exceeds it
  on the weaker Qwen3-1.7B, and stays below the stronger raw baseline on
  Qwen3-8B. DeepSeek-OCR (blue) is competitive on Qwen3-8B at low compression but
  degrades as compression increases, and text summarization (red) is the weakest
  point on every reader. All results use the same 500 questions.}
  \label{fig:longmem-frontier}
\end{figure*}

\paragraph{LongMemEval setup.}
We follow the convention of prior compression
work~\citep{ge2024icae,cheng2024xrag,chevalier2023autocompressor}: train the
compressor on a generic corpus (2{,}000 UltraChat conversations, text only, no
QA labels) and evaluate \emph{zero-shot} on the held-out benchmark. We use the
\emph{oracle} reading setting of LongMemEval~\citep{wu2025longmemeval}, a
test-only benchmark of 500 questions: each question is paired with only its
ground-truth evidence session(s) rather than the full multi-session haystack.
This idealizes the retrieval stage and isolates the reading/representation
problem, which is exactly our scope: we study how history is compressed and
read, not how it is retrieved. We read all provided evidence sessions
untruncated; the longer, distractor-laden LongMemEval-S/M haystacks exceed the
history lengths our compressor is trained on and would confound compression with
retrieval, so we leave pairing LatentPress with a retriever to future work. The
compressor never sees these conversations during training. The reader is a frozen
Qwen2.5-7B-Instruct~\citep{qwen25}. We report overall memory-QA accuracy and the
mean compression ratio (original tokens / compressed vectors), and break out the
\texttt{single-session-user} (precise user fact) and \texttt{knowledge-update}
categories. Baselines: (i) \emph{uncompressed oracle evidence} ($1\times$);
(ii) \emph{uniform soft-token}, our compressor with a single factor over the
whole history; (iii) \emph{text summary}, an LLM-generated summary; (iv)
\emph{DeepSeek-OCR}~\citep{wei2025deepseekocr}, render-to-image visual
compression at three resolutions, run with batched vLLM
inference~\citep{kwon2023vllm}.

\paragraph{Direct soft memory matches uncompressed oracle evidence.}
Table~\ref{tab:main} and Figure~\ref{fig:longmem-frontier} report the comparison on the
Qwen2.5-7B reader. Uncompressed oracle evidence reaches only $0.490$, so more
tokens are not automatically better for this task. In the oracle setting, the
reader receives the correct evidence, but questions can still require
multi-session aggregation, temporal reasoning, knowledge-update tracking, and
abstention. Under this simple role-based schedule, LatentPress reaches $0.476$,
$0.478$, and $0.504$ at $4.62$, $6.27$, and $7.70\times$ compression, matching the
uncompressed reader while using far fewer vectors. A uniform-rate variant of the
same writer, which pools every turn at one rate, stays lower over this range
($0.06$--$0.12$; Table~\ref{tab:main}). The visual baseline is also weaker across the evaluated
curve, decreasing from $0.426$ to $0.312$, and text summary reaches $0.184$.
Because the writer is trained on UltraChat and evaluated on unseen LongMemEval
conversations, this is not benchmark memorization: keeping short user turns
lossless preserves the answer-bearing facts while longer turns are pooled away.

\begin{table*}[t]
  \centering
  \caption{\textbf{Zero-shot LongMemEval} on 500 oracle-evidence questions
  (Llama-3.1-70B-Instruct judge). LatentPress is mean${\pm}$std over five seeds;
  baselines are deterministic. Token-F1 is in Appendix~\ref{app:f1}.}
  \label{tab:main}
  \begin{tabular}{lccc}
    \toprule
    Method & Compression & Overall & user-fact \\
    \midrule
    uncompressed evidence & $1.0\times$ & $0.490$ & $0.946$ \\
    \midrule
    \textbf{LatentPress, $k_a{=}8$}  & $4.62\times$ & $0.476{\pm}0.014$ & $0.938{\pm}0.007$ \\
    \textbf{LatentPress, $k_a{=}16$} & $6.27\times$ & $0.478{\pm}0.020$ & $0.891{\pm}0.015$ \\
    \textbf{LatentPress, $k_a{=}32$} & $7.70\times$ & $\mathbf{0.504{\pm}0.024}$ & $0.938{\pm}0.010$ \\
    \midrule
    ICAE~\citep{ge2024icae} & $4.12\times$  & $0.452{\pm}0.017$ & $0.548{\pm}0.019$ \\
    ICAE~\citep{ge2024icae} & $8.96\times$  & $0.318{\pm}0.022$ & $0.381{\pm}0.023$ \\
    ICAE~\citep{ge2024icae} & $17.28\times$ & $0.174{\pm}0.029$ & $0.209{\pm}0.031$ \\
    \midrule
    DeepSeek-OCR & $2.33\times$ & $0.426$ & $0.797$ \\
    DeepSeek-OCR & $5.97\times$ & $0.390$ & $0.672$ \\
    DeepSeek-OCR & $9.34\times$ & $0.312$ & $0.594$ \\
    \midrule
    text summary & $12.06\times$ & $0.184$ & $0.297$ \\
  \end{tabular}
\end{table*}

\paragraph{The result generalizes across backbones.}
We repeat the zero-shot comparison on Qwen2.5-7B, Qwen3-8B, and Qwen3-1.7B, which
span two model families and a $4.7\times$ range in scale~\citep{qwen25,qwen3},
training one compressor head per reader with the borrowed encoder layers frozen
(see the \texttt{train\_encoder} ablation in Appendix~\ref{sec:ablation}). The
frontier holds on all three: at matched compression LatentPress beats uniform pooling
by $+0.34$ to $+0.45$ in overall accuracy, so reader scale alone does not close
the gap. It also stays close to or ahead of the visual baseline, leading by
$0.504$ vs.\ $0.426$ on Qwen2.5-7B and $0.434$ vs.\ $0.264$ on Qwen3-1.7B, and on
Qwen3-8B trailing only at the lowest compression before overtaking OCR as
compression grows (Figure~\ref{fig:longmem-frontier} and
Table~\ref{tab:crossmodel}). Text summarization stays the weakest baseline on
every reader (per-category breakdown in Appendix~\ref{app:summary_categories}).

\begin{table*}[t]
  \centering
  \caption{\textbf{Zero-shot LongMemEval generalization across Qwen backbones}
  (all 500 questions, UltraChat-trained, Llama-3.1-70B-Instruct judge).
  LatentPress uses $k_a{=}8/16/32$ and reports mean${\pm}$std over five seeds;
  only the reader-specific compressor head is trained, with the borrowed encoder
  layers frozen (Table~\ref{tab:encoder}). Appendix~\ref{app:summary_categories}
  gives the text-summary breakdown.}
  \label{tab:crossmodel}
  \resizebox{\textwidth}{!}{%
  \begin{tabular}{lccccccc}
    \toprule
    & \multicolumn{3}{c}{LatentPress (ours)} & \multicolumn{3}{c}{DeepSeek-OCR} & text summary \\
    \cmidrule(lr){2-4}\cmidrule(lr){5-7}\cmidrule(lr){8-8}
    Reader & $k_a{=}8$ ($4.62\times$) & $k_a{=}16$ ($6.27\times$) & $k_a{=}32$ ($7.70\times$) & $2.33\times$ & $5.97\times$ & $9.34\times$ & \\
    \midrule
    Qwen2.5-7B  & $0.476{\pm}0.014$ & $0.478{\pm}0.020$ & $\mathbf{0.504{\pm}0.024}$ & $0.426$ & $0.390$ & $0.312$ & $0.184$ ($12.1\times$) \\
    Qwen3-8B    & $0.506{\pm}0.015$ & $0.514{\pm}0.020$ & $0.494{\pm}0.025$ & $\mathbf{0.542}$ & $0.506$ & $0.408$ & $0.348$ ($11.3\times$) \\
    Qwen3-1.7B & $\mathbf{0.434{\pm}0.018}$ & $0.424{\pm}0.024$ & $0.416{\pm}0.028$ & $0.264$ & $0.236$ & $0.156$ & $0.106$ ($28.7\times$) \\
    \bottomrule
  \end{tabular}%
  }
\end{table*}

\section{Generalization to Long-Document QA}
\label{sec:longbenchqa}

\begin{figure*}[t]
  \centering
  \includegraphics[width=\linewidth]{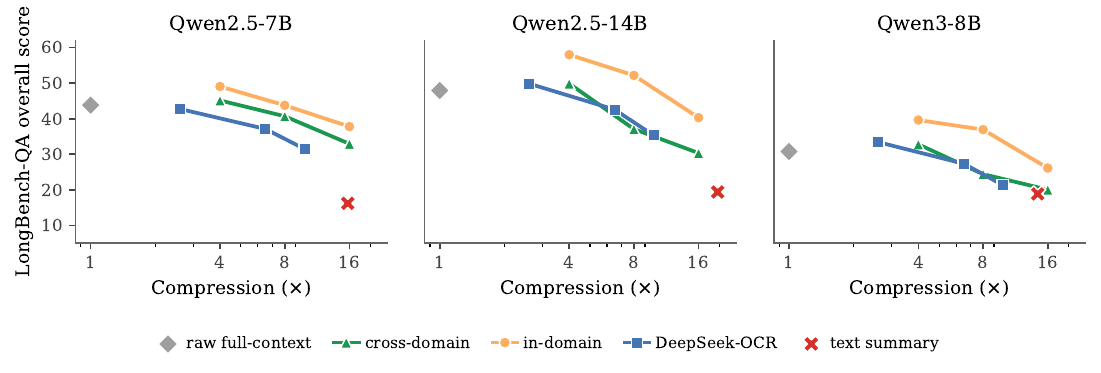}
  \caption{\textbf{LongBench-QA accuracy--compression frontiers.}
  Overall score for Qwen2.5-7B, Qwen2.5-14B, and Qwen3-8B under cross-domain
  (green) and in-domain (orange) writer training. Gray diamonds mark
  uncompressed performance at
  $1\times$, and the blue DeepSeek-OCR curve spans \texttt{base\_size}
  $1024/512$ ($\sim\!2.6/9.9\times$). The red text-summary baseline
  (one point per reader, at $14$--$20\times$) is the weakest on every reader.
  In-domain adaptation exceeds the uncompressed result at the milder rates but
  drops below it at $16\times$ on all three readers.}
  \label{fig:longbench-frontier}
\end{figure*}

Long-document QA tests whether the same interface remains accurate when the
conversational role structure is removed. We therefore use uniform compression
and ask whether cross-domain or in-domain QA supervision can make compressed
soft tokens useful for long-\emph{document} QA on LongBench-QA English~\citep{bai2024longbench}, across six subsets
(\texttt{narrativeqa}, \texttt{qasper}, \texttt{multifieldqa\_en},
\texttt{hotpotqa}, \texttt{2wikimqa}, and \texttt{musique}). We first establish
the uncompressed readers as the reference point: Qwen2.5-14B scores $47.93$,
followed by Qwen2.5-7B at $43.80$ and Qwen3-8B at $30.80$ under its non-thinking
decoding mode (full per-subset results in Table~\ref{tab:lbqa_raw}). We evaluate compressed
configurations on Qwen2.5-7B, Qwen2.5-14B, and
Qwen3-8B. We compare cross-domain training against in-domain task adaptation and
find that compressed readers can match or exceed their own full-context
baselines, although the best rate depends on the reader and task. For Qwen3
readers, raw-context, OCR, and LatentPress runs use the same non-thinking decoding
mode; Appendix~\ref{app:details} gives the exact protocol.

\subsection{Cross-domain QA transfer}
Appendix Table~\ref{tab:lbqa_frontier} asks whether the interface transfers
beyond conversational memory without target-domain training. We sweep the frozen
soft-token compressor at $4$, $8$, and $16\times$ against the raw readers, with
the compressor trained on LongMemEval-derived QA. Transfer is only partly
successful: on Qwen2.5-7B the compressed reader exceeds its own raw baseline at
$4\times$ ($45.13$ vs.\ $43.80$) but falls below it at higher rates ($40.69$ and
$32.94$), and on Qwen3-8B the $4\times$ setting is the single best configuration
($32.79$ vs.\ $30.80$). Across all three readers the mild $4\times$ rate is
preferred and accuracy declines as compression grows; part of Qwen3-8B's drop at
higher rates is a formatting pathology rather than semantic loss, which we
analyze in Appendix~\ref{app:lbqa_failures}.
Thus, cross-domain transfer roughly matches the raw reader at low compression but
does not consistently beat it, which motivates the in-domain adaptation below.

\subsection{In-domain task adaptation}
In-domain task adaptation tests the same accuracy claim in the strongest
task-specific setting. The cross-domain sweep above deliberately transfers from
LongMemEval-derived QA to LongBench-QA with no target-domain training, which
produces unstable per-rate behavior (e.g.\ the $f8$ dip). We therefore train the
same frozen soft-token compressor directly on the LongBench-QA training splits
(NarrativeQA, Qasper, HotpotQA, 2WikiMultihopQA, and MuSiQue) and evaluate on
the matching test subsets, so training and evaluation now share a domain.

The effect is strongest at mild compression: in-domain training lifts the
$4\times$ rate (and, on the larger readers, $8\times$) above the uncompressed
baseline, while the aggressive $16\times$ rate falls below it
(Table~\ref{tab:lbqa_adapt}). Qwen2.5-14B rises from raw $47.93$ to
$57.99$/$52.18$ at $4$/$8\times$ before dropping to $40.30$ at $16\times$;
Qwen2.5-7B beats its raw $43.80$ at $4\times$ ($49.06$), matches it at $8\times$
($43.77$), and falls to $37.78$ at $16\times$; and Qwen3-8B improves over its raw
$30.80$ at $4\times$ and $8\times$ ($39.62$ and $36.93$) but drops to $26.12$ at
$16\times$. As in the cross-domain setting, higher compression tends to help
less, and the most aggressive rate eventually exposes the cost of losing verbatim
detail. The one cost, relative to our zero-shot LongMemEval result, is that this
variant is trained in-domain rather than transferred.
Figure~\ref{fig:longbench-frontier} summarizes the comparison across all three
backbones and compression rates.

\begin{table}[t]
  \centering
  \caption{\textbf{In-domain LatentPress on LongBench-QA English} (official
  overall score in \%, mean${\pm}$std over five seeds). Writers are trained on
  the target-domain training splits; raw rows repeat Table~\ref{tab:lbqa_raw}.}
  \label{tab:lbqa_adapt}
  \footnotesize
  \begin{tabular}{llc}
    \toprule
    Reader & Setting & Overall \\
    \midrule
    \multirow{4}{*}{Qwen2.5-7B}
      & raw context ($1\times$) & $43.80$ \\
      & in-domain $f4$ & $\mathbf{49.06{\pm}2.30}$ \\
      & in-domain $f8$  & $43.77{\pm}2.83$ \\
      & in-domain $f16$ & $37.78{\pm}3.46$\\
    \midrule
  \multirow{4}{*}{Qwen3-8B}
      & raw context ($1\times$) & $30.80$ \\
      & in-domain $f4$  & $\mathbf{39.62{\pm}2.31}$ \\
      & in-domain $f8$  & $36.93{\pm}2.82$ \\
      & in-domain $f16$ & $26.12{\pm}3.33$ \\
    \midrule
    \multirow{4}{*}{Qwen2.5-14B}
      & raw context ($1\times$) & $47.93$ \\
      & in-domain $f4$  & $\mathbf{57.99{\pm}2.35}$ \\
      & in-domain $f8$  & $52.18{\pm}2.82$ \\
      & in-domain $f16$ & $40.30{\pm}3.51$ \\
    \bottomrule
  \end{tabular}
\end{table}

\section{Efficiency}
\label{sec:efficiency}

Having established the accuracy frontier, we separate efficiency into two
deployment costs. The \emph{write cost} is the time to generate encoded tokens;
the \emph{read cost} is the frozen decoder's latency when answering from those
tokens. We also report a coarser end-to-end job time in
Appendix~\ref{app:efficiency_lbqa}.

\paragraph{Write cost.}
Writing soft tokens is a single forward pass, not an autoregressive generation or
OCR-reconstruction process. We measure this encoded-token generation cost on
LongMemEval with a Qwen3-8B backbone in \texttt{bfloat16} on one NVIDIA H100
80GB GPU, using warm-up and synchronized timing. With batches of eight, LatentPress
takes $43$\,ms per conversation. The batched DeepSeek-OCR pipeline renders pages and
reconstructs text by autoregressive optical decoding, taking $844$--$1056$\,ms
per conversation ($\sim\!934$\,ms on average), or about $22\times$ longer at
this stage. Text summarization takes $407$--$645$\,ms, or $9$--$15\times$
longer. The closest soft-token baseline, ICAE~\citep{ge2024icae}, takes
$350$--$700$\,ms per conversation, or $8$--$15\times$ longer, because it encodes
with the full LLM rather than LatentPress's borrowed bottom $L{=}2$ layers; its
read-time latency is comparable to LatentPress, since both inject a short continuous
prefix into the frozen decoder. These speedups are specific to the evaluated
models, output lengths, and batching regimes, and compare LatentPress only with
reconstruction-based routes, not with soft-token methods that also write in one
or a few forward passes.

\paragraph{Read cost.}
For deployment-time latency, we separately measure inference only on $30$
LongBench-QA examples with all models warm-loaded, excluding training, official
evaluation, model loading, and OCR-cache generation (Table~\ref{tab:inference_latency}).
At the $f8$ operating point, LatentPress takes $0.43$--$0.49$ seconds per example,
versus $2.44$--$4.14$ seconds for raw full context and $2.71$--$4.34$ seconds
for cached DeepSeek-OCR at \texttt{base\_size}${=}640$. Across Qwen2.5-7B,
Qwen2.5-14B, and Qwen3-8B, LatentPress is therefore $5.0$--$9.2\times$ faster than
raw inference and $5.5$--$9.4\times$ faster than the cached OCR route.

\begin{table}[t]
  \centering
  \caption{\textbf{Warm-loaded inference-only latency on LongBench-QA}
  (seconds per example; $30$ examples). Times exclude training, official
  evaluation, model loading, and OCR-cache generation. LatentPress uses $f8$;
  DeepSeek-OCR uses a precomputed \texttt{base\_size}${=}640$ cache.}
  \label{tab:inference_latency}
  \footnotesize
  \begin{tabular}{lccc}
    \toprule
    Reader & \makecell{Raw\\context} & \makecell{LatentPress\\$f8$} & \makecell{Cached OCR\\b640} \\
    \midrule
    Qwen2.5-7B  & $2.44$ & $0.49$ & $2.71$ \\
    Qwen2.5-14B & $4.14$ & $0.49$ & $4.34$ \\
    Qwen3-8B    & $3.97$ & $0.43$ & $4.03$ \\
    \bottomrule
  \end{tabular}
\end{table}

Beyond per-conversation writing, the advantage persists end-to-end. On
LongBench-QA, in-domain LatentPress's whole-job time (adapter training, prediction,
and evaluation) is $6.0$--$13.7\times$ shorter than the cold-cache DeepSeek-OCR
pipeline (OCR reconstruction, prediction, and evaluation) at the nearest
available compression settings, and the gap is largest on Qwen2.5-14B
(Appendix~\ref{app:efficiency_lbqa}). Unlike the per-conversation write cost
above (in milliseconds), this is a coarser job-level wall-clock comparison, but
both point the same way.

\section{Related Work}

\paragraph{Soft-token context compression.}
A line of work compresses context into a few continuous vectors that a decoder
consumes in place of text. Gist~\citep{mu2023gist},
AutoCompressor~\citep{chevalier2023autocompressor}, and ICAE~\citep{ge2024icae}
adapt an LLM-scale reader or encoder; xRAG~\citep{cheng2024xrag} projects one
retrieved passage into a single token, and 500xCompressor~\citep{li2024500x}
targets still higher ratios. LatentPress instead compresses multi-turn histories
and whole documents end-to-end with no retriever, while keeping the reader frozen
and training only a reader-matched adapter. Its vectors enter the input-embedding
layer directly, require no reconstruction, and can use different rates across
structured segments (Table~\ref{tab:positioning}). We compare with ICAE under the
same frozen reader and LongMemEval evaluation (Table~\ref{tab:main}). Context
compression also benefits production retrieval and ranking systems
\citep{behdin2025scaling}, though that setting differs from conversational memory.

LCLMs~\citep{li2026endtoendcontextcompressionscale} scale direct latent-context
compression through joint encoder--decoder training, whereas LatentPress keeps
the reader frozen and trains only a small adapter.

\paragraph{Token pruning and visual compression.}
Selective Context~\citep{li2023selective} and the LLMLingua family
\citep{jiang2023llmlingua,jiang2023longllmlingua,pan2024llmlingua2} drop or
select prompt tokens. TIP~\citep{xu2026tip} similarly shows that token-level
training value is highly nonuniform. Visual alternatives render text as images:
DeepSeek-OCR~\citep{wei2025deepseekocr} reconstructs text optically, while
Glyph~\citep{cheng2025glyph} and AgentOCR~\citep{feng2025agentocr} apply visual
compression to long context or agent histories. Reconstruction-based visual
compression saves context tokens but incurs an autoregressive OCR stage before a
language-only reader can answer. LatentPress instead writes vectors that the same
frozen decoder reads directly. On sparse conversational memory, we find that the
OCR route degrades monotonically with compression (Section~\ref{sec:longmemeval}).

\paragraph{Conversational memory systems.}
Systems such as Generative Agents~\citep{park2023generative},
MemoryBank~\citep{zhong2024memorybank}, MemGPT~\citep{packer2023memgpt}, and
Mem0~\citep{chhikara2025mem0} manage long histories with retrieval, reflection,
or summarization pipelines and are evaluated on benchmarks like
LongMemEval~\citep{wu2025longmemeval}. They typically store and retrieve text;
LatentPress is complementary, providing a learned soft-token representation that
could sit inside such pipelines.

\paragraph{Latent reasoning.}
Compressing a model's own reasoning into continuous states, as in
Coconut~\citep{hao2024coconut}, and shortening reasoning traces through
self-distillation, as in CRISP~\citep{sang2026crisp}, are related in spirit but
target the generation trace rather than the input history. LatentPress never alters
what the reader generates, only what it reads.

\section{Conclusion and Future Work}
\label{sec:future}

LatentPress shows that compressed context need not be stored as text or reconstructed
from images before a language model can use it. A small reader-matched writer
maps conversational histories and long documents into continuous memory tokens
that a frozen decoder reads directly through its embedding interface. Across
LongMemEval and LongBench-QA, this interface satisfies the four practical
criteria motivating it: compressed readers preserve or improve accuracy, encoded
tokens are generated in near-real time, reading from the compressed prefix is much
faster than reading raw context or the cached OCR route, and the trainable state is
only a small adapter while the decoder remains frozen. These results establish
soft tokens as a practical machine-facing context interface beyond text and
vision, with dynamic compression as a natural next step for pushing compression
further.

A central direction for future work is dynamic compression. The role-based and
uniform rates used here are deliberately simple, hand-specified heuristics;
instead of fixing them, a learned policy could choose the compression rate per
segment, preserving detail only where it matters and compressing the rest more
aggressively. Such a policy could be optimized with reinforcement learning
against downstream answer reward under a latency or memory budget, building on
learned prompt-compression and token-importance
signals~\citep{xu2026tip,jiang2023llmlingua,pan2024llmlingua2,li2023selective}.
The general formulation also permits a learned token-wise fusion $H$ of literal
and contextual features. Another natural extension is to train writers for
additional readers and for non-text context such as tool, multimodal, or
embodied traces.

\section*{Limitations}

LongMemEval uses oracle evidence and excludes distractor-heavy LongMemEval-S/M.
Compression rates are hand-specified, and the role-based schedule may not
transfer beyond user-centric memory. Each reader requires its own writer. On
LongBench-QA, gains require in-domain adaptation at mild rates, while performance
degrades at $16\times$. Experiments cover only English text and Qwen-family
readers, and timing results use one H100. Finally, soft tokens are less
interpretable and harder to debug than text.

\bibliography{references}

@inproceedings{mu2023gist,
  title={Learning to compress prompts with gist tokens},
  author={Mu, Jesse and Li, Xiang Lisa and Goodman, Noah},
  booktitle={Advances in Neural Information Processing Systems (NeurIPS)},
  year={2023}
}

@inproceedings{ge2024icae,
  title={In-context autoencoder for context compression in a large language model},
  author={Ge, Tao and Hu, Jing and Wang, Lei and Wang, Xun and Chen, Si-Qing and Wei, Furu},
  booktitle={International Conference on Learning Representations (ICLR)},
  year={2024}
}

@inproceedings{chevalier2023autocompressor,
  title={Adapting language models to compress contexts},
  author={Chevalier, Alexis and Wettig, Alexander and Ajith, Anirudh and Chen, Danqi},
  booktitle={Conference on Empirical Methods in Natural Language Processing (EMNLP)},
  year={2023}
}

@article{xu2026tip,
  title={{TIP}: Token Importance in On-Policy Distillation},
  author={Xu, Yuanda and Sang, Hejian and Zhou, Zhengze and He, Ran and Wang, Zhipeng and Geramifard, Alborz},
  journal={arXiv preprint arXiv:2604.14084},
  year={2026}
}

@inproceedings{cheng2024xrag,
  title={{xRAG}: Extreme context compression for retrieval-augmented generation with one token},
  author={Cheng, Xin and Wang, Xun and Zhang, Xingxing and Ge, Tao and Chen, Si-Qing and Wei, Furu and Zhang, Huishuai and Zhao, Dongyan},
  booktitle={Advances in Neural Information Processing Systems (NeurIPS)},
  year={2024}
}

@article{sang2026crisp,
  title={{CRISP}: Compressed Reasoning via Iterative Self-Policy Distillation},
  author={Sang, Hejian and Xu, Yuanda and Zhou, Zhengze and He, Ran and Wang, Zhipeng and Sun, Jiachen},
  journal={arXiv preprint arXiv:2603.05433},
  year={2026}
}

@article{li2024500x,
  title={500xCompressor: Generalized prompt compression for large language models},
  author={Li, Zongqian and Su, Yixuan and Collier, Nigel},
  journal={arXiv preprint arXiv:2408.03094},
  year={2024}
}

@inproceedings{jiang2023llmlingua,
  title={{LLMLingua}: Compressing prompts for accelerated inference of large language models},
  author={Jiang, Huiqiang and Wu, Qianhui and Lin, Chin-Yew and Yang, Yuqing and Qiu, Lili},
  booktitle={Conference on Empirical Methods in Natural Language Processing (EMNLP)},
  year={2023}
}

@inproceedings{jiang2023longllmlingua,
  title={{LongLLMLingua}: Accelerating and enhancing {LLMs} in long context scenarios via prompt compression},
  author={Jiang, Huiqiang and Wu, Qianhui and Luo, Xufang and Li, Dongsheng and Lin, Chin-Yew and Yang, Yuqing and Qiu, Lili},
  booktitle={Annual Meeting of the Association for Computational Linguistics (ACL)},
  year={2024}
}

@inproceedings{pan2024llmlingua2,
  title={{LLMLingua-2}: Data distillation for efficient and faithful task-agnostic prompt compression},
  author={Pan, Zhuoshi and Wu, Qianhui and Jiang, Huiqiang and Xia, Menglin and Luo, Xufang and Zhang, Jue and Lin, Qingwei and R{\"u}hle, Victor and Yang, Yuqing and Qiu, Lili and others},
  booktitle={Findings of the Association for Computational Linguistics (ACL)},
  year={2024}
}

@article{li2023selective,
  title={Unlocking context constraints of {LLMs}: Enhancing context efficiency of {LLMs} with self-information-based content filtering},
  author={Li, Yucheng},
  journal={arXiv preprint arXiv:2304.12102},
  year={2023}
}

@article{behdin2025scaling,
  title={Scaling Up Efficient Small Language Models Serving and Deployment for Semantic Job Search},
  author={Behdin, Kayhan and Song, Qingquan and Vasudevan, Sriram and Sheng, Jian and Ma, Xiaojing and Zhou, Z. and Zhu, Chuanrui and Li, Guoyao and Nguyen, Chanh and others},
  journal={arXiv preprint arXiv:2510.22101},
  year={2025}
}

@article{wei2025deepseekocr,
  title={{DeepSeek-OCR}: Contexts optical compression},
  author={Wei, Haoran and Sun, Yaofeng and Li, Yukun},
  journal={arXiv preprint arXiv:2510.18234},
  year={2025}
}

@article{cheng2025glyph,
  title={Glyph: Scaling context windows via visual-text compression},
  author={Cheng, Jiale and Liu, Yusen and Zhang, Xinyu and Fei, Yulin and Hong, Wenyi and others},
  journal={arXiv preprint arXiv:2510.17800},
  year={2025}
}

@article{feng2025agentocr,
  title={{AgentOCR}: Reimagining Agent History via Optical Self-Compression},
  author={Feng, Lang and Yang, Fuchao and Chen, Feng and Cheng, Xin and Xu, Haiyang and Wan, Zhenglin and Yan, Ming and An, Bo},
  journal={arXiv preprint arXiv:2601.04786},
  year={2026}
}

@article{wu2025longmemeval,
  title={{LongMemEval}: Benchmarking chat assistants on long-term interactive memory},
  author={Wu, Di and Wang, Hongwei and Yu, Wenhao and Zhang, Yuwei and Chang, Kai-Wei and Yu, Dong},
  journal={International Conference on Learning Representations (ICLR)},
  year={2025}
}

@article{ding2023ultrachat,
  title={Enhancing chat language models by scaling high-quality instructional conversations},
  author={Ding, Ning and Chen, Yulin and Xu, Bokai and Qin, Yujia and Zheng, Zhi and Hu, Shengding and Liu, Zhiyuan and Sun, Maosong and Zhou, Bowen},
  journal={Conference on Empirical Methods in Natural Language Processing (EMNLP)},
  year={2023}
}

@article{qwen25,
  title={Qwen2.5 technical report},
  author={Yang, An and Yang, Baosong and Zhang, Beichen and others},
  journal={arXiv preprint arXiv:2412.15115},
  year={2024}
}

@article{qwen3,
  title={Qwen3 technical report},
  author={Yang, An and Li, Anfeng and Yang, Baosong and others},
  journal={arXiv preprint arXiv:2505.09388},
  year={2025}
}

@article{srivastava2015highway,
  title={Highway networks},
  author={Srivastava, Rupesh Kumar and Greff, Klaus and Schmidhuber, J{\"u}rgen},
  journal={arXiv preprint arXiv:1505.00387},
  year={2015}
}

@inproceedings{cho2014gru,
  title={Learning phrase representations using {RNN} encoder-decoder for statistical machine translation},
  author={Cho, Kyunghyun and van Merri{\"e}nboer, Bart and Gulcehre, Caglar and Bahdanau, Dzmitry and Bougares, Fethi and Schwenk, Holger and Bengio, Yoshua},
  booktitle={Proceedings of the 2014 Conference on Empirical Methods in Natural Language Processing (EMNLP)},
  year={2014}
}

@inproceedings{bai2024longbench,
  title={{LongBench}: A bilingual, multitask benchmark for long context understanding},
  author={Bai, Yushi and Lv, Xin and Zhang, Jiajie and Lyu, Hongchang and Tang, Jiankai and Huang, Zhidian and Du, Zhengxiao and Liu, Xiao and Zeng, Aohan and Hou, Lei and Dong, Yuxiao and Tang, Jie and Li, Juanzi},
  booktitle={Annual Meeting of the Association for Computational Linguistics (ACL)},
  year={2024}
}

@article{packer2023memgpt,
  title={{MemGPT}: Towards {LLM}s as operating systems},
  author={Packer, Charles and Wooders, Sarah and Lin, Kevin and Fang, Vivian and Patil, Shishir G and Stoica, Ion and Gonzalez, Joseph E},
  journal={arXiv preprint arXiv:2310.08560},
  year={2023}
}

@inproceedings{park2023generative,
  title={Generative agents: Interactive simulacra of human behavior},
  author={Park, Joon Sung and O'Brien, Joseph C. and Cai, Carrie J. and Morris, Meredith Ringel and Liang, Percy and Bernstein, Michael S.},
  booktitle={ACM Symposium on User Interface Software and Technology (UIST)},
  year={2023}
}

@article{zhong2024memorybank,
  title={{MemoryBank}: Enhancing large language models with long-term memory},
  author={Zhong, Wanjun and Guo, Lianghong and Gao, Qiqi and Ye, He and Wang, Yanlin},
  journal={Proceedings of the AAAI Conference on Artificial Intelligence},
  year={2024}
}

@article{chhikara2025mem0,
  title={{Mem0}: Building production-ready {AI} agents with scalable long-term memory},
  author={Chhikara, Prateek and Khant, Dev and Aryan, Saket and Singh, Taranjeet and Yadav, Deshraj},
  journal={arXiv preprint arXiv:2504.19413},
  year={2025}
}

@inproceedings{hao2024coconut,
  title={Training large language models to reason in a continuous latent space},
  author={Hao, Shibo and Sukhbaatar, Sainbayar and Su, DiJia and Li, Xian and Hu, Zhiting and Weston, Jason and Tian, Yuandong},
  booktitle={Conference on Language Modeling (COLM)},
  year={2025}
}

@inproceedings{kwon2023vllm,
  title={Efficient memory management for large language model serving with {PagedAttention}},
  author={Kwon, Woosuk and Li, Zhuohan and Zhuang, Siyuan and Sheng, Ying and Zheng, Lianmin and Yu, Cody Hao and Gonzalez, Joseph E and Zhang, Hao and Stoica, Ion},
  booktitle={Symposium on Operating Systems Principles (SOSP)},
  year={2023}
}

@article{dubey2024llama,
  title={The Llama 3 Herd of Models},
  author={Dubey, Abhimanyu and Jauhri, Abhinav and Pandey, Abhinav and others},
  journal={arXiv preprint arXiv:2407.21783},
  year={2024}
}

@misc{li2026endtoendcontextcompressionscale,
  title={End-to-End Context Compression at Scale},
  author={Ang Li and Sean McLeish and Haozhe Chen and Nimit Kalra and Zaiqian Chen and Artem Gazizov and Venkata Anoop Suhas Kumar Morisetty and Bhavya Kailkhura and Harshitha Menon and Zhuang Liu and Brian R. Bartoldson and Tom Goldstein and Sanae Lotfi and Micah Goldblum and Pavel Izmailov},
  year={2026},
  eprint={2606.09659},
  archivePrefix={arXiv},
  primaryClass={cs.CL},
  url={https://arxiv.org/abs/2606.09659}
}

\appendix

\section{Implementation and Training Details}
\label{app:details}

\subsection{Architecture}
The compressor reuses the bottom $L{=}2$ transformer layers of the frozen
Qwen2.5-7B-Instruct decoder~\citep{qwen25} as an encoder; these layers are
deep-copied so that gradient updates do not perturb the reader. A small trainable
head on top of this encoder maps its features into soft tokens: a linear adapter
$A\in\mathbb{R}^{d\times d}$ (with $d$ the model hidden size) initialized to the
identity, so the writer begins close to the raw token embeddings and departs from
them only as training warrants. Only this head is
trained: $12.849$M parameters for
Qwen2.5-7B, $16.781$M for Qwen3-8B, $4.196$M for Qwen3-1.7B, and $26.220$M for
Qwen2.5-14B. The two borrowed encoder layers and all decoder
weights are frozen (we deep-copy the borrowed layers only to hold a stable,
reader-matched encoder, not to update them; see the \texttt{train\_encoder}
ablation, Table~\ref{tab:encoder}). Soft tokens are injected through the decoder's
input-embedding interface (\texttt{inputs\_embeds}), so the reader's forward pass
is unchanged.

\subsection{Training Objective}
Equation~\ref{eq:training} gives the full objective. Both losses are averaged
over non-padding target positions; padding is masked from the reconstruction and
forward-KL terms.

\subsection{Hyperparameters}
Table~\ref{tab:hparams} lists the full configuration. Training data are token
chunks drawn from UltraChat conversations: a conversation shorter than the chunk
length is used whole (padded, with padding masked out of both the reconstruction
and forward-KL losses), and a longer conversation is split into windows. The
reader, tokenizer, and optimizer are shared across all runs. Generalization runs
(Table~\ref{tab:crossmodel}) retrain the compressor head against each reader
(Qwen3-8B, Qwen3-1.7B) with all other settings unchanged.

\begin{table*}[t]
  \centering
  \caption{Training and model hyperparameters. The uniform baseline uses a single
  factor $k\in\{4,8,16\}$; the role-aware model uses $k_{\text{user}}{=}1$
  (lossless) and $k_{\text{assistant}}\in\{8,16,32\}$.}
  \label{tab:hparams}
  \resizebox{\textwidth}{!}{%
  \begin{tabular}{lc}
    \toprule
    Hyperparameter & Value \\
    \midrule
    Decoder (reader, frozen) & Qwen2.5-7B / Qwen2.5-14B / Qwen3-8B / Qwen3-1.7B (frozen)~\citep{qwen25,qwen3} \\
    Encoder layers $L$ (borrowed, \textbf{frozen}) & $2$ \\
    Writer head & linear adapter $d\times d$, identity-initialized (trained) \\
    Trainable parameters & backbone-dependent (4.196M--26.220M; 12.849M for Qwen2.5-7B) \\
    \midrule
    Optimizer & AdamW \\
    Learning rate & $1\times10^{-4}$ \\
    Training steps & $1000$ \\
    Number of training chunks & $400$ \\
    Chunking & short conversations padded (mask-aware); long ones windowed \\
    Chunk length (\texttt{max\_len}) & $2048$ tokens \\
    Batch size & $1$ \\
    \midrule
    Reconstruction loss & teacher-forced cross-entropy (pad-masked) \\
    Forward-KL weight $\lambda$ & $1.0$ (per-token mean, pad-masked) \\
    Precision & bf16 (decoder), fp32 (compressor head) \\
    \midrule
    Uniform pooling factor $k$ & $\{4, 8, 16\}$ \\
    Role-aware $k_{\text{user}}$ / $k_{\text{assistant}}$ & $1$ / $\{8,16,32\}$ \\
    \midrule
    Training corpus & UltraChat~\citep{ding2023ultrachat}, 2{,}000 conversations, zero-shot \\
    Evaluation & LongMemEval~\citep{wu2025longmemeval}, 500 questions \\
    Eval sampling & \texttt{shuffle}, seed $0$, greedy decoding (temperature $0$) \\
    \midrule
    Judge model & Llama-3.1-70B-Instruct~\citep{dubey2024llama} (official LongMemEval model zoo) \\
    Judge serving & vLLM, FP8, tensor-parallel $2$ \\
    Judge decoding & temperature $0$, \texttt{max\_tokens}${=}10$, verdict ${=}$ ``yes'' in output \\
    Judge prompt & official per-question-type templates (no system prompt) \\
    \bottomrule
  \end{tabular}%
  }
\end{table*}

\subsection{Decoding}
All answers are generated deterministically with greedy decoding
(\texttt{do\_sample}${=}$\texttt{False}, i.e.\ temperature $0$), so results are
reproducible and free of sampling variance. The soft-token reader generates up to
$64$ new tokens per answer; the visual-baseline reader generates up to $256$.
DeepSeek-OCR's text reconstruction step likewise uses greedy vLLM decoding
(temperature $0$, up to $2048$ tokens per rendered page).

\subsection{Compression Ratio}
For a conversation, the compression ratio is the number of original history
tokens divided by the number of injected vectors,
$\rho = |\mathcal{C}| / \sum_i \lceil n_i / k_{r_i}\rceil$ with $n_i$ the length
of turn $i$. In the role-aware model, lossless user turns ($k_{\text{user}}{=}1$)
contribute one vector per token, so $\rho$ is driven by the assistant rate and
the user/assistant token mix; the reported ratios ($4.62$--$7.70\times$) are
means over the evaluation set rather than a preset budget.

\section{Evaluation and Baseline Details}
\label{app:baselines}

\subsection{LongBench-QA Cross-Domain Setup}
For the cross-domain LongBench-QA sweep, we keep the same writer architecture
and training recipe and train reader-specific weights
on LongMemEval-style QA triples, a separate supervision regime from the
UltraChat training used for LongMemEval evaluation. All cross-domain runs use
\texttt{configs/simple.json} with \texttt{pool=mean},
\texttt{qa\_train=true}, and \texttt{train\_encoder=false}; the decoder is
frozen, including the borrowed encoder layers, and only the compressor head is
trained. Training data are \texttt{longmemeval\_qa\_train.json};
the compression factors are $4$, $8$, and $16$; and evaluation is on the six
LongBench-QA English subsets listed in Table~\ref{tab:lbqa_raw}. Raw baselines
use the same evaluation split but skip soft-token training entirely.

\subsection{LongBench-QA In-Domain Setup}
For in-domain adaptation, we retain the same reader-matched writer, uniform mean
pooling, and compression factors $4$, $8$, and $16$, but replace the
LongMemEval-derived supervision with target-domain QA examples. The training
pool combines the training splits of NarrativeQA, Qasper, HotpotQA,
2WikiMultihopQA, and MuSiQue; evaluation uses the six LongBench-QA English
subsets listed in Table~\ref{tab:lbqa_raw}, including MultiFieldQA-en. We train a
separate writer for each frozen Qwen2.5-7B, Qwen2.5-14B, and Qwen3-8B reader.
The decoder and its two borrowed encoder layers remain frozen, so only the
reader-specific compressor head is optimized. Each run uses $1{,}000$ updates,
batch size $1$ with gradient accumulation $8$, a bf16 decoder, and an fp32
compressor head. Qwen3-8B uses non-thinking decoding in all conditions. We score
predictions with the official LongBench-QA prompt templates and \texttt{eval.py}
F1 implementation; Table~\ref{tab:lbqa_adapt} reports mean${\pm}$standard
deviation over five training seeds.

\subsection{Text-Summary Baseline}
The text-summary baseline replaces the history with an LLM-generated abstractive
summary that the same frozen reader then answers from. The summarizer is prompted
with a fixed system instruction (``You compress conversations into a dense
factual summary that preserves every concrete fact either speaker stated about
themselves (names, dates, preferences, events, relationships, numbers). Omit
small talk. Be terse.'') and a user instruction ``Compress the following
conversation into at most \{budget\} tokens, preserving all concrete personal
facts,'' followed by the conversation text. The token budget adapts to the input
length, $\mathrm{budget}=\max(128,\ \lfloor n_{\mathrm{tok}}/r_{\mathrm{sum}}\rfloor)$,
where $n_{\mathrm{tok}}$ is the conversation length and $r_{\mathrm{sum}}$ the
target summary ratio.

\subsection{DeepSeek-OCR on LongMemEval}
The visual baseline renders each conversation to page images (fixed render size
$1024{\times}1024$, font size $16$) and reconstructs text with
DeepSeek-OCR~\citep{wei2025deepseekocr} at its three officially recommended
resolution modes
(\texttt{base\_size} $\in\{1024, 640, 512\}$, giving $256/100/64$ vision tokens
per page and mean compression $2.33/5.97/9.34\times$ on LongMemEval). The
reconstructed text is then read by the same frozen reader, matching the
soft-token answering protocol and the same 500-question set. We use DeepSeek-OCR
in its native reconstruction mode: it is an optical \emph{OCR} model whose
reported metric is text-decoding precision~\citep{wei2025deepseekocr}, so
recovering text and reading it with a language model is its intended
compression-then-read use, unlike Glyph~\citep{cheng2025glyph}, which renders
text and answers \emph{directly} with a VLM. Reconstructing text also lets the
\emph{same} frozen reader answer for both LatentPress and the visual baseline, which
isolates the compression representation from the reader rather than confounding
it with a different (vision-language) model. Reconstruction uses
batched vLLM inference~\citep{kwon2023vllm} (v0.11.2,
\texttt{DeepseekOCRForCausalLM}), processing all rendered pages in a single
\texttt{generate} call; this reduced reconstruction time from an estimated
$\sim\!15$ hours (sequential transformer decoding) to a few minutes. At the
highest $1024$-pixel resolution, vLLM's CUDA-graph capture triggered an
illegal-memory-access error on the full 500-conversation batch, so that setting
was run with \texttt{enforce\_eager=True}.

\subsection{DeepSeek-OCR on LongBench-QA}
The LongBench-QA visual baseline uses the same two-stage pipeline as the
LongMemEval one above, and differs only where the benchmark requires it. The
LongBench context is plain text rather than PDF pages, so we first render it to
page images at a fixed \texttt{render\_size}${=}1024$ and
\texttt{font\_size}${=}18$; each page is then resized to \texttt{base\_size}
$\in\{512,640,1024\}$ and passed to DeepSeek-OCR with the prompt
\texttt{<image>\textbackslash nFree OCR.} to reconstruct the text. The
reconstructed text is inserted into the official LongBench-QA prompt template in
place of the raw context, answered by the same frozen Qwen reader, and scored
with the official LongBench-QA \texttt{eval.py} (F1), identical to the raw and
soft-token runs. As on LongMemEval, the OCR stage only reconstructs text; the
question is answered by the text-only decoder, not by the vision model. We count
vision tokens with the implementation's \texttt{tokens\_per\_page(base\_size)},
which yields $73/111/273$ tokens per page for
\texttt{base\_size}${=}512/640/1024$, slightly above the official $64/100/256$
because of extra query/special-token overhead, so our
\texttt{b512}/\texttt{b640}/\texttt{b1024} sweep follows the same frontier as the
official Tiny/Small/Base modes with marginally higher effective token counts. At
\texttt{base\_size}${=}640$ this corresponds to about $6.5\times$ effective
text-to-vision-token compression on LongBench-QA.

\section{Additional LongMemEval Results}

\subsection{Disentangling Role Allocation and Writer Learning}
\label{app:role_writer_ablation}

Table~\ref{tab:role_writer_ablation} separates the contribution of the
role-based allocation rule from that of the learned writer. All variants use
the same frozen Qwen2.5-7B reader, $k_a{=}8$, and evaluation protocol. The no-learning control keeps
user turns unchanged but replaces the learned assistant representation with
directly pooled embeddings; the user-only control removes assistant turns; and
the role-swapped control uses $k_u{=}8$ and reverses which role is preserved
verbatim.

\begin{table*}[t]
  \centering
  \caption{\textbf{Role-allocation and writer ablation on LongMemEval}
  (Qwen2.5-7B, $k_a{=}8$).
  Compression is measured as original history tokens divided by injected
  vectors. Results use the same 500-question oracle-evidence evaluation as the
  main LongMemEval experiments.}
  \label{tab:role_writer_ablation}
  \begin{tabular}{lcccc}
    \toprule
    Setting & User turns & Assistant turns & Compression & Overall \\
    \midrule
    Full LatentPress & raw embeddings & learned soft tokens & $4.62\times$ & $0.476{\pm}0.014$ \\
    No-learning & raw embeddings & pooled embeddings & $4.62\times$ & $0.325{\pm}0.023$ \\
    User-only & raw embeddings & removed & $9.6\times$ & $0.217{\pm}0.031$ \\
    Role-swapped & learned soft tokens & raw embeddings & $1.10\times$ & $0.087{\pm}0.011$ \\
    \bottomrule
  \end{tabular}
\end{table*}

The ablation separates the gains from representation learning and role-based
allocation. At the same $4.62\times$ compression, replacing the learned writer
with direct embedding pooling reduces accuracy from $0.476$ to $0.325$, showing
that the result does not come from the role schedule alone. Removing assistant
turns further lowers accuracy to $0.217$, so even the pooled assistant context
retains information complementary to the lossless user turns. Finally, reversing
the allocation---compressing user turns while preserving assistant turns---drops
accuracy to $0.087$. This confirms that answer-bearing information in
LongMemEval is concentrated in user turns and supports preserving them verbatim
under the current hand-specified schedule. More generally, the large sensitivity
to this allocation motivates learning when to write and how much to compress,
for example with a reinforcement-learning policy optimized for downstream
accuracy under a memory or latency budget.

\subsection{Encoder-Training Ablation}
\label{sec:ablation}
The soft-token writer borrows the reader's bottom two transformer layers to
encode context before pooling. We can either fine-tune those layers alongside the
compressor head (\texttt{train\_encoder}${=}$true) or freeze them and train
only the head (\texttt{train\_encoder}${=}$false). Table~\ref{tab:encoder}
compares the two on Qwen2.5-7B under the identical recipe. Freezing the encoder
wins at every rate ($+0.018$ to $+0.066$ absolute accuracy) and, unlike
fine-tuning, does not degrade at high compression. Fine-tuning on UltraChat
appears to overfit the training distribution, whereas the frozen layers retain
the reader's general representation. We therefore use
\texttt{train\_encoder}${=}$false throughout.

\begin{table*}[t]
  \centering
  \caption{\textbf{Encoder-training ablation} (LongMemEval, Qwen2.5-7B,
  Llama-3.1-70B judge). Freezing the borrowed encoder layers and training only the
  compressor head beats fine-tuning them at every assistant rate.}
  \label{tab:encoder}
  \begin{tabular}{lccc}
    \toprule
    Borrowed encoder & $k_a{=}8$ & $k_a{=}16$ & $k_a{=}32$ \\
    \midrule
    fine-tuned (\texttt{train\_encoder}=true)  & $0.454$ & $0.460$ & $0.438$ \\
    \textbf{frozen (\texttt{train\_encoder}=false)} & $\mathbf{0.476}$ & $\mathbf{0.478}$ & $\mathbf{0.504}$ \\
    \bottomrule
  \end{tabular}
\end{table*}


\subsection{Judge-Free Token-F1}
\label{app:f1}

Our primary metric follows the official LongMemEval protocol: per-question-type
LLM-judge accuracy, which we score with Llama-3.1-70B-Instruct, one of the
judges in the official LongMemEval model zoo (Appendix~\ref{app:details}). To
confirm that the ranking is not an artifact of the judge, we also
report a deterministic, judge-free metric: token-level F1 between the model
response and the gold answer, computed on the same
500 LongMemEval runs as Table~\ref{tab:main}. We truncate each response at the
first hallucinated follow-up turn before scoring, identically to the accuracy
protocol.

Absolute F1 is low for all methods because the readers emit explanatory
sentences (e.g.\ ``You attended the Maundy Thursday service at the Episcopal
Church'') while gold answers are short spans (``the Episcopal Church''), which
dilutes precision. The \emph{ordering}, however, is identical to judge-accuracy:
role-aware LatentPress dominates the uniform rate by $3$--$4\times$ and is similar to
DeepSeek-OCR at the lowest reported operating points and higher thereafter, with
the gap largest at high compression ($k_a{=}32$: $0.251$ vs.\ DeepSeek-OCR
$0.160$). Two independent
metrics (LLM-judge accuracy and token-F1) thus agree on the ranking.

\begin{table*}[t]
  \centering
  \caption{\textbf{Token-level F1 on LongMemEval} (Qwen2.5-7B reader, same 500
  runs as Table~\ref{tab:main}). Judge-free auxiliary metric; the ordering
  matches judge-accuracy.}
  \label{tab:f1}
  \begin{tabular}{lcc}
    \toprule
    Method & Compression & token-F1 \\
    \midrule
    \textbf{role-aware (ours), $k_a{=}8$}  & $4.62\times$ & $0.230$ \\
    \textbf{role-aware (ours), $k_a{=}16$} & $6.27\times$ & $0.222$ \\
    \textbf{role-aware (ours), $k_a{=}32$} & $7.70\times$ & $\mathbf{0.251}$ \\
    \midrule
    uniform soft-token, $k{=}4$   & $4.00\times$  & $0.072$ \\
    uniform soft-token, $k{=}8$   & $7.99\times$  & $0.052$ \\
    uniform soft-token, $k{=}16$  & $15.96\times$ & $0.059$ \\
    \midrule
    DeepSeek-OCR (visual) & $2.33\times$ & $0.233$ \\
    DeepSeek-OCR (visual) & $5.97\times$ & $0.208$ \\
    DeepSeek-OCR (visual) & $9.34\times$ & $0.160$ \\
    \bottomrule
  \end{tabular}
\end{table*}

\subsection{Text-Summary Per-Category Breakdown}
\label{app:summary_categories}

Table~\ref{tab:summary_categories} breaks down the text-summary baseline by
LongMemEval question type, scored with the same official Llama-3.1-70B-Instruct
judge and per-type protocol as the main results (Appendix~\ref{app:details}).
The overall column matches the summary rows used in
Section~\ref{sec:longmemeval}. The breakdown explains why text summarization is
the weakest baseline on every reader: abstention accuracy is high (the reader
correctly declines when a fact is absent), but the answer-bearing categories
collapse: \texttt{temporal} and \texttt{multi-session} fall to $0.016$--$0.041$
on the smaller readers because abstractive summarization discards the precise
user facts that these questions require. Qwen2.5-14B is included for reference
only; it is a LongBench-QA reader and is not used elsewhere on LongMemEval.

\begin{table*}[t]
  \centering
  \caption{\textbf{Text-summary baseline, per-category accuracy on LongMemEval}
  (Llama-3.1-70B judge, official per-type protocol, 500 questions). Overall and
  achieved compression are reported alongside the seven question-type
  accuracies. Qwen2.5-14B is shown for reference (a LongBench-QA reader).}
  \label{tab:summary_categories}
  \resizebox{\textwidth}{!}{%
  \begin{tabular}{lccccccccc}
    \toprule
    Reader & Overall & Comp. & abstention & knowledge-update & multi-session & single-assistant & single-preference & single-user & temporal \\
    \midrule
    Qwen2.5-7B  & $0.184$ & $12.06\times$ & $0.967$ & $0.111$ & $0.017$ & $0.554$ & $0.033$ & $0.297$ & $0.016$ \\
    Qwen3-8B    & $0.348$ & $11.29\times$ & $0.767$ & $0.292$ & $0.207$ & $0.464$ & $0.433$ & $0.641$ & $0.197$ \\
    Qwen3-1.7B  & $0.106$ & $28.73\times$ & $0.300$ & $0.069$ & $0.041$ & $0.250$ & $0.100$ & $0.109$ & $0.079$ \\
    \midrule
    Qwen2.5-14B$^\dagger$ & $0.352$ & $16.48\times$ & $0.933$ & $0.361$ & $0.207$ & $0.625$ & $0.500$ & $0.516$ & $0.110$ \\
    \bottomrule
  \end{tabular}%
  }
  \\[2pt]
  {\footnotesize $^\dagger$Reference only; Qwen2.5-14B is a LongBench-QA reader.}
\end{table*}

\subsection{Qualitative Examples}
\label{app:examples}

Table~\ref{tab:examples} shows representative LongMemEval questions together with
the gold answer and the answer produced by our role-aware compressor
(\textbf{ours}, $k_a{=}8$, Qwen3-8B reader) reading \emph{only} the compressed
history. Each example lists the raw$\rightarrow$compressed token counts. Despite
$4$--$6\times$ compression, precise user facts (names, durations, prices, days of
the week) survive because user turns are kept lossless; the model also correctly
\emph{abstains} when the queried fact was never stated.

\begin{table*}[t]
  \centering
  \caption{\textbf{Qualitative examples on LongMemEval} (ours, $k_a{=}8$,
  Qwen3-8B reader, read from the compressed history only). Predictions are
  lightly truncated for space.}
  \label{tab:examples}
  \small
  \begin{tabular}{p{2.4cm}p{4.3cm}p{1.6cm}p{2.8cm}}
    \toprule
    Category (compression) & Question & Gold & Ours \\
    \midrule
    temporal-reasoning \newline ($5.2\times$) &
    Who did I meet with during the lunch last Tuesday? &
    Emma & Emma \checkmark \\
    \addlinespace
    knowledge-update \newline ($4.5\times$) &
    What day of the week do I take a cocktail-making class? &
    Friday & Friday \checkmark \\
    \addlinespace
    multi-session \newline ($4.2\times$) &
    How much more expensive was the taxi ride compared to the train fare? &
    \$6 & ``\$6 more than the train fare'' \checkmark \\
    \addlinespace
    single-session-user \newline ($5.0\times$) &
    How long have I been collecting vintage cameras? &
    three months & 3 months \checkmark \\
    \addlinespace
    single-session-\newline assistant ($4.7\times$) &
    Remind me of the romantic Italian restaurant in Rome you recommended? &
    Roscioli & Roscioli \checkmark \\
    \addlinespace
    abstention \newline ($3.5\times$) &
    Which did I start first, the Ferrari model or the Porsche 991 Turbo S model? &
    Not enough information (Porsche never mentioned) &
    ``NOT MENTIONED\ldots{} only the Ferrari'' \checkmark \\
    \bottomrule
  \end{tabular}
\end{table*}

\section{Additional LongBench-QA Analysis}

\subsection{Raw LongBench-QA Results}

\begin{table*}[t]
  \centering
  \caption{\textbf{Raw LongBench-QA English baseline} (overall and per-subset
  scores). These are the uncompressed references for the cross-domain and
  in-domain sweeps.}
  \label{tab:lbqa_raw}
  \resizebox{\textwidth}{!}{%
  \begin{tabular}{lccccccc}
    \toprule
    Reader & Overall & narrativeqa & qasper & multifieldqa\_en & hotpotqa & 2wikimqa & musique \\
    \midrule
    Qwen2.5-7B & 43.80 & 29.29 & 44.14 & 52.32 & 58.40 & 47.80 & 30.85 \\
    Qwen2.5-14B & 47.93 & 29.89 & 45.23 & 53.85 & 62.23 & 58.86 & 37.52 \\
    Qwen3-8B   & 30.80 & 16.01 & 31.36 & 41.67 & 39.32 & 34.87 & 21.56 \\
    \bottomrule
  \end{tabular}%
  }
\end{table*}

\subsection{Cross-Domain Transfer Results}

\begin{table}[t]
  \centering
  \caption{\textbf{LatentPress on LongBench-QA English} (frozen soft-token
  compressor, official overall scores in \%). Compression factors $f4/f8/f16$
  correspond to $3.999/7.997/15.985\times$. The compressor is trained on
  LongMemEval-derived QA (cross-domain transfer), and the best operating point is
  reader-dependent. Overall is reported as mean${\pm}$std over $5$ training seeds.
  Raw rows repeat the uncompressed reference from Table~\ref{tab:lbqa_raw}.}
  \label{tab:lbqa_frontier}
  \footnotesize
  \begin{tabular}{llc}
    \toprule
    Reader & Setting & Overall \\
    \midrule
    \multirow{4}{*}{Qwen2.5-7B}
      & raw context ($1\times$)          & $43.80$ \\
      & cross-domain $f4$ ($4\times$)   & $45.13{\pm}2.41$ \\
      & cross-domain $f8$ ($8\times$)   & $40.69{\pm}2.20$ \\
      & cross-domain $f16$ ($16\times$) & $32.94{\pm}3.24$ \\
    \midrule
    \multirow{4}{*}{Qwen3-8B}
      & raw context ($1\times$)          & $30.80$ \\
      & cross-domain $f4$ ($4\times$)   & $\mathbf{32.79{\pm}1.29}$ \\
      & cross-domain $f8$ ($8\times$)   & $24.41{\pm}1.67$ \\
      & cross-domain $f16$ ($16\times$) & $20.05{\pm}2.01$ \\
    \midrule
    \multirow{4}{*}{Qwen2.5-14B}
      & raw context ($1\times$)          & $47.93$ \\
      & cross-domain $f4$ ($4\times$)   & $49.88{\pm}1.28$ \\
      & cross-domain $f8$ ($8\times$)   & $37.08{\pm}1.65$ \\
      & cross-domain $f16$ ($16\times$) & $30.34{\pm}2.27$ \\
    \bottomrule
  \end{tabular}
\end{table}

\subsection{Failure Modes}
\label{app:lbqa_failures}

The aggregate LongBench-QA scores in Section~\ref{sec:longbenchqa} hide several
qualitatively distinct failure modes that emerge as compression increases.
Table~\ref{tab:lbqa_failures} collects representative cases. These are
degradations of the \emph{generated content and format}, not artifacts of the
evaluation harness: the JSONL schema and scoring are identical to the raw runs.
We observe six recurring patterns: (i) \emph{unanswerable collapse}, where the
compressed reader abstains on a question it answers from full context; (ii)
\emph{format artifacts}, where the reader wraps its answer in a JSON envelope;
(iii) \emph{blank output}; (iv) \emph{repetition loops}, most common at
$16\times$; (v) \emph{reasoning-template leakage}, where a Qwen3
\texttt{</think>} tag escapes into the answer; and (vi) \emph{well-formed but
semantically wrong} short answers, where the output has the right style but the
wrong fact. The first five are decoding or formatting pathologies that become
more common as compression increases; the last reflects genuine information loss
in the compressed memory.

\begin{table*}[t]
  \centering
  \caption{\textbf{LongBench-QA failure modes} under the frozen soft-token
  compressor. Gold answers are abbreviated; outputs are verbatim (lightly
  truncated). These are content/format degradations, not evaluation-harness
  issues.}
  \label{tab:lbqa_failures}
  \small
  \begin{tabular}{p{2.2cm}p{1.7cm}p{3.0cm}p{3.7cm}}
    \toprule
    Failure mode & Reader ($f$) & Gold (abbrev.) & Compressed output \\
    \midrule
    Unanswerable collapse & Qwen2.5-7B ($f8$, $f16$) &
    ``extension of NetVLAD, adds Ghost clusters\ldots'' &
    \texttt{unanswerable} \\
    \addlinespace
    Format artifact (JSON) & Qwen3-8B ($f8$) &
    ``ground truth is not established in the paper'' &
    \texttt{unanswerable \{"answer": "unanswerable"\}} \\
    \addlinespace
    Blank output & Qwen3-8B ($f16$) &
    ``guest in the home of the Mulvilles'' &
    \emph{(empty string)} \\
    \addlinespace
    Repetition loop & Qwen3-8B ($f16$) &
    ``I have seen the Lord.'' &
    \texttt{Answer: Answer: Answer: \ldots} \\
    \addlinespace
    \texttt{</think>} leak & Qwen3-8B ($f8$) &
    ``Watt, one joule per second.'' &
    \texttt{Watt </think>\{"answer": "Watt"\}} \\
    \addlinespace
    Well-formed but wrong & Qwen2.5-7B ($f8$, $f16$) &
    ``guest in the home of the Mulvilles'' &
    ``Homeless'' / ``Homeless on the streets.'' \\
    \bottomrule
  \end{tabular}
\end{table*}

\section{End-to-End Efficiency Details}
\label{app:efficiency_lbqa}

\subsection{LongBench-QA Wall-Clock Comparison}

Table~\ref{tab:lbqa_speed} reports whole-job wall-clock time on the LongBench-QA
English evaluation for in-domain LatentPress and the DeepSeek-OCR visual baseline at
the nearest available compression settings. This is a coarser, job-level view
than the per-conversation write latency in the main text (reported in
milliseconds). Each job ran once on one NVIDIA H100 80GB GPU, so these are
single-run wall-clock measurements without variance estimates. LatentPress includes
adapter training, prediction, and official evaluation. Training uses batch size
$1$, gradient accumulation $8$, $1{,}000$ steps, a bf16 decoder, and a frozen
reader; prediction processes one LongBench example at a time before official
\texttt{eval.py} scoring. DeepSeek-OCR cold-cache time includes reconstruction,
reader prediction, and evaluation. Reconstruction uses vLLM with
\texttt{max\_num\_seqs}${=}16$, \texttt{max\_num\_batched\_tokens}${=}16384$,
and \texttt{gpu\_memory\_utilization}${=}0.70$; reader prediction again processes
one example at a time.

Under this cold-cache accounting, LatentPress is $6.0$--$13.7\times$ faster, with the
largest gap between Qwen2.5-14B \texttt{b512} ($9.9\times$) and the nearest
LatentPress $f8$ point ($7.997\times$). OCR caches
depend on the rendered context and resolution but not on the downstream reader,
so one cache can be reused across readers. Table~\ref{tab:lbqa_speed_amortized}
therefore also reports an amortized four-reader scenario in which cache-generation
time is divided by four. LatentPress remains faster under this favorable OCR
accounting. Qwen3-8B OCR used a single \texttt{stage=both} job, so its cache and
reader times cannot be separated and are reported only as cold-cache totals.

\begin{table*}[t]
  \centering
  \caption{\textbf{Cold-cache end-to-end wall-clock time on LongBench-QA
  English} (minutes), measured once on one H100 80GB GPU. We report the total
  time to go from raw context to a scored prediction, broken into stages.
  \emph{Cache} is the one-time cost of building the method's intermediate
  artifact before any question is answered: for DeepSeek-OCR this is rendering
  the context to images and reconstructing text by optical decoding; LatentPress has
  no such stage (\emph{---}), since it trains a small adapter instead of
  precomputing a cache. \emph{Pred.+eval} is reader prediction plus official
  \texttt{eval.py} scoring. \emph{Cold total} is the whole-job time from scratch:
  for LatentPress, adapter training${+}$prediction${+}$evaluation; for DeepSeek-OCR,
  cache generation${+}$prediction${+}$evaluation (i.e.\ the cost when no cache
  exists yet, hence ``cold''). \emph{Score} is the official overall F1 (\%) at
  that operating point; the LatentPress scores match Table~\ref{tab:lbqa_adapt},
  while these DeepSeek-OCR resolutions (\texttt{b512}/\texttt{b1024}) are reported
  only here. Qwen3-8B DeepSeek-OCR
  ran cache and prediction in one \texttt{stage=both} job, so its per-stage times
  are merged (\emph{incl.}) and only the cold total is available. Points are
  nearest available compression settings, not exact matches.}
  \label{tab:lbqa_speed}
  \resizebox{\textwidth}{!}{%
  \begin{tabular}{lllccccc}
    \toprule
    Reader & Method & Point & Compression & Cache & Pred.+eval & Cold total & Score \\
    \midrule
    \multirow{5}{*}{Qwen2.5-7B}
      & in-domain LatentPress & $f4$   & $3.999\times$  & --- & --- & 18.6  & 49.06 \\
      & in-domain LatentPress & $f8$   & $7.997\times$  & --- & --- & 15.4  & 43.77 \\
      & in-domain LatentPress & $f16$  & $15.985\times$ & --- & --- & 14.0  & 37.78 \\
      & DeepSeek-OCR & b512  & $9.9\times$ & 102.9 & 64.9 & 167.8 & 31.48 \\
      & DeepSeek-OCR & b1024 & $2.6\times$ & 90.5  & 50.0 & 140.5 & 42.75 \\
    \midrule
    \multirow{5}{*}{Qwen2.5-14B}
      & in-domain LatentPress & $f4$   & $3.999\times$  & --- & --- & 25.1  & 57.99 \\
      & in-domain LatentPress & $f8$   & $7.997\times$  & --- & --- & 20.8  & 52.18 \\
      & in-domain LatentPress & $f16$  & $15.985\times$ & --- & --- & 21.2  & 40.30 \\
      & DeepSeek-OCR & b512  & $9.9\times$ & 102.9 & 182.7 & 285.6 & 35.29 \\
      & DeepSeek-OCR & b1024 & $2.6\times$ & 90.5  & 68.2  & 158.7 & 49.86 \\
    \midrule
    \multirow{5}{*}{Qwen3-8B}
      & in-domain LatentPress & $f4$   & $3.999\times$  & --- & --- & 23.0  & 39.62 \\
      & in-domain LatentPress & $f8$   & $7.997\times$  & --- & --- & 20.6  & 36.93 \\
      & in-domain LatentPress & $f16$  & $15.985\times$ & --- & --- & 18.2  & 26.12 \\
      & DeepSeek-OCR & b512  & $9.9\times$ & incl. & incl. & 165.4 & 21.31 \\
      & DeepSeek-OCR & b1024 & $2.6\times$ & incl. & incl. & 137.9 & 33.41 \\
    \bottomrule
  \end{tabular}
  }
\end{table*}
\subsection{OCR Cache Amortization}
\begin{table*}[t]
  \centering
  \caption{\textbf{DeepSeek-OCR cache amortization across four readers}
  (minutes). Amortized time is reader prediction/evaluation plus one quarter of
  the reader-independent cache-generation time. Qwen3-8B is omitted because its
  stage times were not logged separately.}
  \label{tab:lbqa_speed_amortized}
  \begin{tabular}{lcccc}
    \toprule
    Reader and OCR point & OCR compression & Cold total & Amortized total & Nearest LatentPress point \\
    \midrule
    Qwen2.5-7B, b512   & $9.9\times$ & 167.8 & 90.6  & $f8$ ($7.997\times$): 15.4 \\
    Qwen2.5-7B, b1024  & $2.6\times$ & 140.5 & 72.6  & $f4$ ($3.999\times$): 18.6 \\
    Qwen2.5-14B, b512  & $9.9\times$ & 285.6 & 208.4 & $f8$ ($7.997\times$): 20.8 \\
    Qwen2.5-14B, b1024 & $2.6\times$ & 158.7 & 90.8  & $f4$ ($3.999\times$): 25.1 \\
    \bottomrule
  \end{tabular}
\end{table*}
\end{document}